\documentclass[10pt,twocolumn,letterpaper]{article}

\usepackage{wacv}              

\usepackage{graphicx}
\usepackage{amsmath}
\usepackage{amssymb}
\usepackage{booktabs}
\usepackage{microtype}      
\usepackage{xcolor}         
\usepackage{algorithm}
\usepackage{algorithmic}
\usepackage{wrapfig}

\usepackage{amsmath}

\usepackage[numbers]{natbib}
\usepackage{euscript}

\newenvironment{packed_itemize}
{\begin{itemize}
    \setlength{\itemsep}{1pt}
    \setlength{\parskip}{0pt}
    \setlength{\parsep}{0pt}
}{\end{itemize}}
\usepackage[pagebackref,breaklinks,colorlinks]{hyperref}

\newcommand{\shortname}{EgoSonics\xspace}
\newcommand{\modulename}{SyncroNet\xspace}
\title{\shortname: Generating Synchronized Audio for Silent Egocentric Videos}

\usepackage[capitalize]{cleveref}
\crefname{section}{Sec.}{Secs.}
\Crefname{section}{Section}{Sections}
\Crefname{table}{Table}{Tables}
\crefname{table}{Tab.}{Tabs.}

\def\wacvPaperID{489} 
\def\confName{WACV}
\def\confYear{2025}

\author{
Aashish Rai
\and
Srinath Sridhar
\vspace{0.01in}
\and
\centerline{Brown University}
\and
{\tt\small \url{https://ivl.cs.brown.edu/research/egosonics}}
}

\begin{document}
\twocolumn[{%
\renewcommand\twocolumn[1][]{#1}%
\vspace{-1.0cm}
\maketitle
\vspace{-0.8cm}
\centering
    \includegraphics[width=0.98\textwidth]{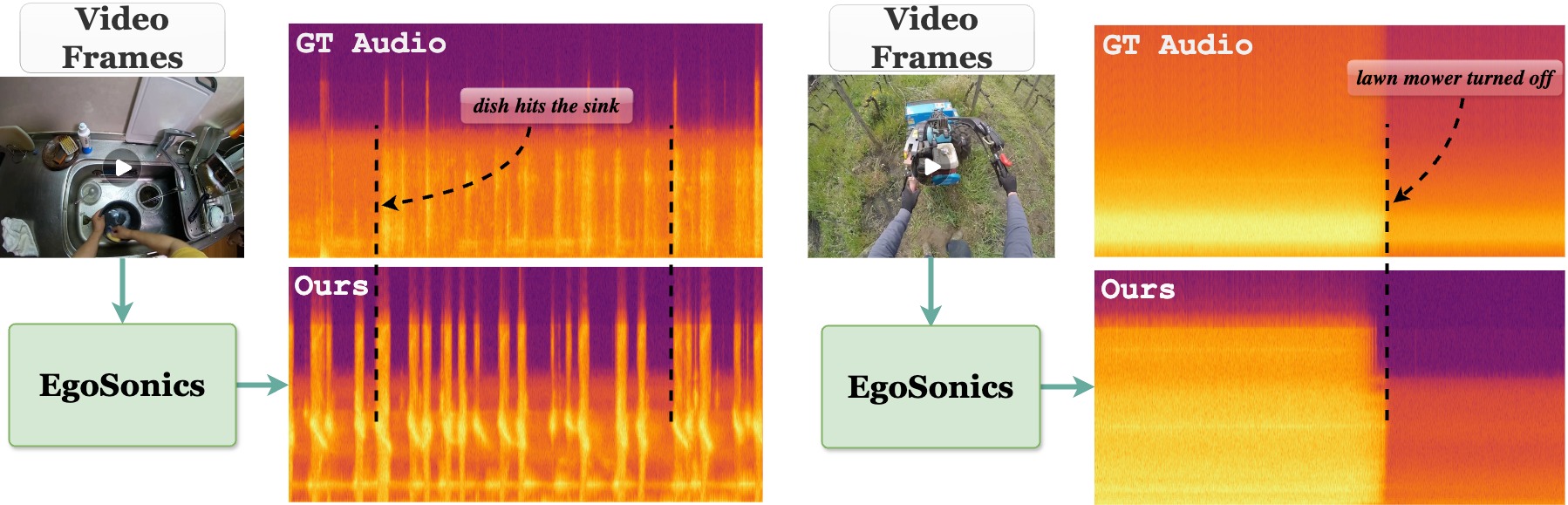}
    \captionof{figure}{\textit{
    We present \textbf{\shortname}, a method to synthesize audio tracks conditioned on silent in-the-wild egocentric videos. 
    Our method operate on videos at 30~fps, and can synthesize audio that is semantically meaningful and synchronized with events in the video (``dish hits the sink'' or ``lawn mower turned off'').
    We also propose a new method to evaluate audio-video synchronization quality.
    }}
    \vspace*{0.3cm}
    \label{fig:results_1}
}]

%

\vspace{-1cm}

\begin{abstract}
\vspace{-3mm}
We introduce \emph{\shortname}, a method to generate semantically meaningful and synchronized audio tracks conditioned on silent egocentric videos.
Generating audio for silent egocentric videos could open new applications in virtual reality, assistive technologies, or for augmenting existing datasets.
Existing work has been limited to domains like speech, music, or impact sounds and cannot capture the broad range of audio frequencies found in egocentric videos.
\shortname addresses these limitations by building on the strengths of latent diffusion models for conditioned audio synthesis.
We first encode and process paired audio--video data to make them suitable for generation.
The encoded data is then used to train a model that can generate an audio track that captures the semantics of the input video.
Our proposed \textit{\modulename} builds on top of ControlNet to provide control signals that enables generation of temporally-synchronized audio.
%
%
Extensive evaluations and a comprehensive user study show that our model outperforms existing work in audio quality, and in our proposed synchronization evaluation method.
Furthermore, we demonstrate downstream applications of our model in improving video summarization.
%
\end{abstract}

\vspace{-6mm}
\section{Introduction}
\label{sec:intro}
\vspace{-2mm}
As humans, we have the ability to watch a silent video and imagine the sounds that could be accompanying it.
We use visual cues, such as the swaying of branches or the flowing of water, to conjure up the sounds of a tree or river.
Machines with such capabilities could have potentially transformational applications in the movie industry, virtual reality, gaming, and in assisting people with disabilities.
Although we now have models that can generate photorealistic videos from text~\cite{sora,veo,gupta2023photorealistic,makeavideo}, these models cannot yet synthesize corresponding synchronized audio.

Recently, we have seen a surge in multimodal models~\cite{chen2017deep, fanzeres2021sound, hao2018cmcgan, li2022learning, lee2022sound, diff_foley} 
that can learn mappings between domains such as text-to-images, image-to-text, text-to-video, and text-to-audio~\cite{multimodal_survey1, multimodal_survey2}.
Multimodal models that operate with audio have thus far been primarily limited to image-to-audio (I2A)~\cite{sheffer2023hear, iashin2021taming} or audio-to-image (A2I)~\cite{sung2023sound} synthesis.
There are relatively fewer works that explore the video-to-audio (V2A) or audio-to-video (A2V) problems due to the complex temporal and spatial understanding required.
Thus, these works have been limited to domains 
like speech~\cite{choi2023diffv2s, hong2022visagesyntalk, chen2023novel}, music~\cite{su2020audeo}, or impact sounds~\cite{su2023physics}.
Furthermore, existing models have short video contexts (1--4 frames~\cite{make_an_audio, sheffer2023hear, diff_foley, clipsonic}),
cannot produce audio that is synchronized with the input video, and are limited to audio frequencies of less than 8~KHz~\cite{diff_foley, sheffer2023hear, make_an_audio}.

In this paper, we address the problem of \textbf{generating synchronized audio tracks} for silent everyday 
videos.
Specifically, we focus on \emph{egocentric} videos -- videos that are captured using head- or body-worn cameras providing a first-person viewpoint.
Our focus on egocentric videos is driven by potential applications arising from the gaining popularity of wearable and assistive technologies~\cite{soundingactions, avcorrespondance_ego4d, meta_aria, ego_vlp}.
For instance, wearable virtual reality devices can generate immersive 3D visual environments, but audio and sounds are harder to synthesize~\cite{vraudio1,soundingactions}.
Existing egocentric datasets~\cite{ego4d,epic} often lack audio due to privacy concerns, technical constraints, or other design factors.
Even if video generative models~\cite{sora, veo, gupta2023photorealistic} can synthesize very realistic egocentric videos, they cannot yet generate the corresponding synchronized sounds, thus limiting their applications.

We present \textbf{\shortname}, a method that can generate synchronized audio tracks conditioned on silent egocentric videos.
%
We build upon the state-of-the-art generation capabilities of Latent Diffusion Models~\cite{stable_diffusion, ddim, ddpm} to generate realistic audio tracks that are not only semantically meaningful to the visual content of videos but also synchronized to events in them (see \Cref{fig:results_1}).
We cast audio generation as an image generation problem by operating on the Short-Time Fourier Transform ``image''~\cite{griffin_lim_faststft} of raw audio signals that are time-aligned with the input video.
We introduce the \textbf{\modulename} module which builds upon ControlNet~\cite{controlnet} to consume videos in the form of temporally-stacked images.
The use of temporally-aligned and encoded audio-video pairs enables \modulename to learn correspondences through self- and cross-attention. 
We then use Stable Diffusion
model~\cite{stable_diffusion} to generate audio spectrograms conditioned on the control signals from \textit{\modulename} and the video embedding.
%
%
%

Different from previous works, \shortname operates on videos captured at 30~frames per second (fps) to accurately synchronize the synthesized audio, allowing us to effectively capture auditory events in the video.
Unlike previous work that are often limited to audio frequencies of less than 8~KHz, we operate on a broader set of frequencies
(upto 20~KHz) -- an important factor to consider when dealing with egocentric day-to-day activities that contain useful frequencies of up to 16~KHz.
To improve quality even further, we use a learned audio upsampler that super-resolves the generated audio. 
Our method achieves state-of-the-art results in generating high quality, semantically meaningful, and synchronized audio on the Ego4D dataset~\cite{ego4d}.


Measuring synchronization between the synthesized audio and the input video is a challenging problem with no standardized metric in the community.
Therefore, we provide a new way to accurately measure audio-visual (AV) synchronization by extracting AV features through a Vision Transformer (ViT) and training an MLP to learn the alignment between these features in a contrastive manner.
We outperform existing methods including Im2Wav~\cite{sheffer2023hear}, Diff-Foley~\cite{diff_foley}, and Make-an-audio~\cite{make_an_audio} in audio quality and synchronization.
In addition, we also show \shortname can improve video summarization results on egocentric data.
To sum up our main contributions:
\begin{packed_itemize}
    \item We propose \textbf{\shortname}, a method to generate semantically meaningful and synchronized audio for everyday silent egocentric videos.
    \item We propose the \textbf{\modulename} module 
    that can extract the temporal information from a video at 30~fps and produce control signals to enable conditional audio generation with better synchronization.
    \item To overcome the limitations of existing audio-video synchronization metrics, we propose a new method to evaluate synchronization accuracy.
    %
\end{packed_itemize}

We highly encourage readers to see the accompanying videos to fully appreciate our results.






\vspace{-1mm}
\section{Related Works}
\label{gen_inst}

\vspace{-1mm}


\paragraph{Generative Models and Multimodal Learning.}
Diffusion Models \cite{ddim, ddpm, stable_diffusion} have demonstrated remarkable efficacy in a multitude of generative tasks, spanning image generation, audio synthesis, and video creation. 
Among these, latent diffusion models like Stable Diffusion (SD) \cite{stable_diffusion} stand out for their ability to perform both forward and reverse processes within the latent space of data, leading to efficient computation and expedited inference. 
Notably, SD is engineered to execute inference even on standard personal laptop GPU, making it a practical choice for a wide array of users. 
Alongside diffusion models, other generative architectures such as Variational Autoencoders (VAEs) \cite{vahdat2020nvae, cemgil2020autoencoding}, Generative Adversarial Networks (GANs) \cite{gan2020, stylegan2022, rai2021improved, albedogan2024, 3dfacecam2023}, and Autoregressive models have also made significant strides in various domains, each with their own strengths and uses.


With the recent developments in the generative models, we have seen a significant rise in multimodal learning. 
Text guided synthesis has seen a rapid development in the last few years. 
The state-of-the-are DALL-E~\cite{dalle,dalle3} can generate realistic images given natural language text queries by encoding images into latent tokens. Other recent text-to-image (T2I) models include \cite{t2i1,t2i2,t2i3,t2i4}. 
The progress has also been made in text-to-video, where papers like Make-a-Video~\cite{makeavideo} uses text-to-image to sample an image and then uses the temporal understanding of real-world videos to translate the sampled image into a video. Very recently, we have also seen realistic text-to-video synthesis from SORA, VEO~\cite{sora, veo}.
Text-to-audio has also seen some progress in the past few years. However, due to the lack of large amount of text-to-audio datasets, we don't see as much progress as in T2I. Works like \cite{text2audio1, text2music1, music_controlnet}, use datasets like AudioSet~\cite{audioset} for training and are able to generate music from text.

\vspace{-4mm}

\paragraph{Audio-Visual Learning.}
The exploration of audio-visual cross-modal generation encompasses two primary directions: vision-to-sound and sound-to-vision generation. Vision-to-sound tasks have been studied in contexts such as instrument/music and speech generation \cite{chen2017deep, hao2018cmcgan, su2020audeo}. 
Recent efforts by Luo et al. \cite{diff_foley}, Iashin et al. \cite{iashin2021taming}, and Huang et al.~\cite{make_an_audio} aimed to broaden the scope by conditioning sound generation on videos from diverse categories. Majumder et al.~\cite{avcorrespondance_ego4d} tries to learn AV correspondance for active speaker detection and spatial audio denoising. However, our approach surpasses previous limitations, enabling the generation of plausible and synchronized audios from a wide range of everyday activities, including but not limited to \textit{cooking, laundry, carpentry, running, placing items}, etc. 
Unlike earlier methods, our model produces high-quality audio closely related to the input audio. 
Existing V2A methods struggle with synchronization as they lack audio-related information in pretrained visual features, limiting the capture of intricate audio-visual correlations.

Im2Wav \cite{sheffer2023hear} generates an audio given an image. 
It uses a two-transfer model and uses CLIP~\cite{clip} image embeddings are the conditional signal to guide the low-level and up-level transformers. 
Make-An-Audio \cite{make_an_audio} is a text-to-audio generation model that leverages a prompt-enhanced diffusion model. 
It uses a pair of audio encoder-decoder network to convert the audio spectrogram into the rich feature space where the diffusion process is applied. 
Cross-attention is applied with the text embeddings coming from a text encoder. 
It can generate I2A as they leveraged the fact that CLIP image and text embeddings share the same latent space.
For V2A, the model samples 4 frames from the video, uses a pooling layer to get an average embedding and then use it to condition the audio generation. 

Both the Im2Wav \cite{sheffer2023hear} and Make-an-audio \cite{make_an_audio} lacks synchronization in generated audio, as neither single image nor four frames are enough to get semantic and contextual information from the activity being performed in the video and generate the synchronized audio. 
These methods may work for videos with less difference between different frames, eg., \textit{video of an ocean}, \textit{a car on the road}, \textit{musical instruments}, etc. 
But they are not suitable for activity videos where even two consecutive frames might differ and contain rich semantic changes.
For video condition sampling, Im2Wav takes the mean of all the video frame embeddings, and uses it to guide the audio generation. 
However, this loses the temporal understanding of the video and can only give global guidance for audio generation.


The closest paper to ours is Diff-Foley \cite{diff_foley}, which tries to synthesize synchronized audio given a muted video using the a contrastive audio-visual pretraining (CAVP) framework. CAVP tries to align the audio and video features using contrastive learning by bringing similar samples closer and vice-versa. The CAVP-aligned video features are later used to train a latent diffusion model to generate audio. However, the quality of audio generated by the framework is sub-optimal - a result of using lower audio sampling rate and smaller mel basis to fit more data into a smaller spectrogram. Also, in-the-wild testing of the framework on day-to-day activities generate out-of-context and non-synchronized audios. It only uses four frames per second in a video, thereby limiting the amount of information that can be captured from a video. 



\begin{figure*}[!t]
\centering
\includegraphics[width=6.4in]{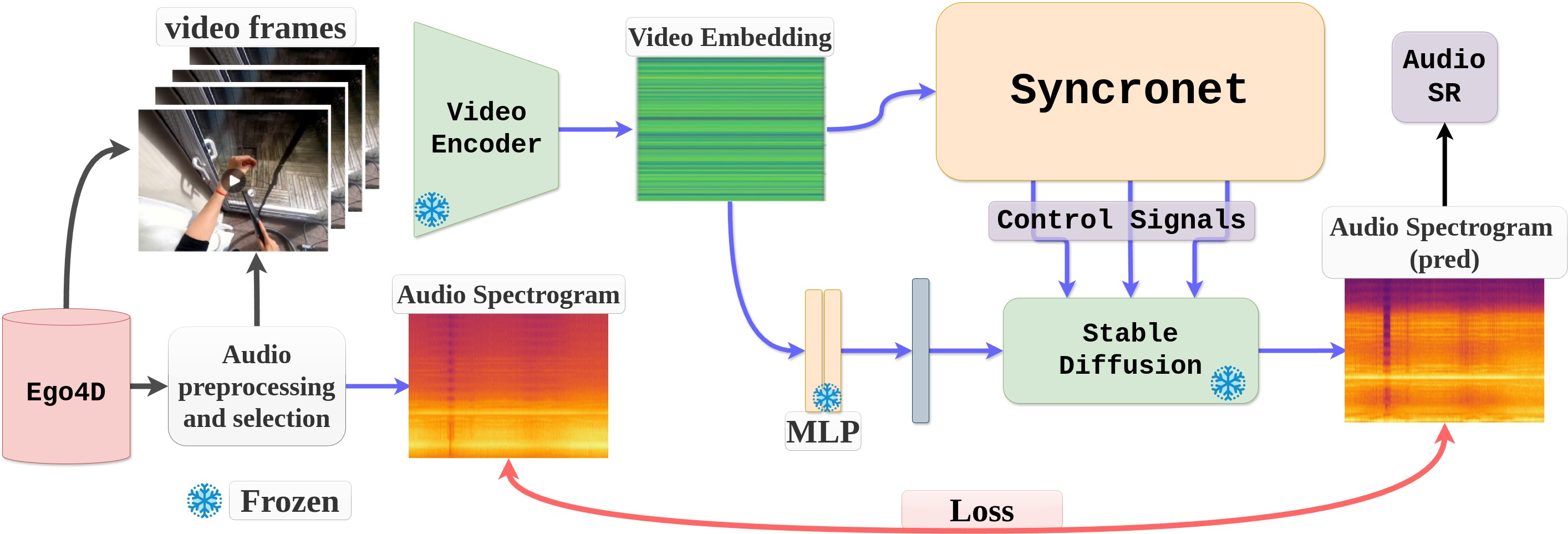} 
\caption{The overall architecture of our proposed method - \shortname.
The input video frames are encoded through a video encoder to get video embedding $E_V$.
This video embedding goes to the \emph{\modulename}~\ref{subsec:syncronet} which generates several control signals to control the generation of audio spectrograms by providing pixel-level temporal control to a pre-trained Stable Diffusion (SD)~\ref{subsec:ldm}. An MLP translates the video embedding into text embedding $c_t$ for SD. The loss between the ground truth audio spectrogram $E_A$ and predicted $E'_A$ is used to train the \modulename.
Finally, as a post processing step, generated audio is upsampled using Audio SR module.}
\label{fig:architecture}
\vspace{-4mm}
\end{figure*}

\vspace{-4mm}

\section{Method}
\label{sec:method}
Given an audio-video pair $(A, V)$, we want to learn a conditional generative model to synthesize audio waveforms $A \in \mathbb{R}^{T' . f_{sa}}$, conditioned on a video $V \in \mathbb{R}^{T' . f_{sv} \times W \times H \times 3 }$, where, $T'$ is the duration of clip in seconds, $f_{sa}$ is the audio sampling rate (generally 16~KHz--48~KHz), $f_{sv}$ is the frame rate of video (generally 30, 45, 60, 90~fps), $W \times H \times 3$ is the shape of RGB video frame. 
Learning a conditional model to generate a high-dimensional audio waveform of size $T\times f_{sa}$ is non-trivial and cumbersome~\cite{diff_foley, sheffer2023hear, text2audio1}, so we first encode the audio ($E_A$) and video ($E_V$) for efficient learning.

Our goal then is to train a conditional generative model that can generate audio encodings $E_A$ given the video embedding $E_V$ with per-frame synchronization - $ \EuScript{P} (E_A | E_V)$. 
\Cref{fig:architecture} shows our approach that consists of a two-step process, where the first stage generates control signals to guide the conditional generative model $ \EuScript{P} (E_A | E_V, C)$, where $C := \{ c^{(n)} \} _{n=1}^N $ are the control signals. 
The control signals provide pixel-level local guidance to a diffusion-based generative model.
We describe each component of our architecture below.
%



\subsection{Audio/Video Preprocessing}
\label{subsec:avprep}

\paragraph{Audio Encoding and Pruning.}
To compactly encode audio, we use the widely-adopted using Short-Time Fourier Transform (STFT)~\cite{griffin_lim_faststft} representation.
A spectrogram, $E_A \in \mathbb{R}^{T \times D}$ is an ``image'' of the audio frequencies over time, obtained through the Fourier transform of the audio signal.
The X-axis of audio spectrograms generally represents time, while the Y-axis represents frequencies at each time step. 
Since we cast audio as spectrogram images, we can use the image generation capabilities of diffusion models to generate audio.

Large datasets like Ego4D~\cite{ego4d} may contain videos where audio is from background activities (e.g.,~traffic) rather than foreground visual cues.
To avoid overfitting to such uninformative audio signals, we only select useful and relevant clips by picking non-overlapping video clips based on a threshold Root Mean Square (RMS) value of the audio waveform.
The audio waveforms $A \in \mathbb{R}^{T' . f_{sa}}$ are converted to spectrograms $E_A \in \mathbb{R}^{T \times D}$.
More details about audio preprocessing and selection are given in the supplementary.

\vspace{-2mm}
\paragraph{Video Encoder.}
To encode videos, we repurpose the CLIP~\cite{clip} image encoder which is trained on a large dataset of images and corresponding text descriptions.
CLIP allowed us to generate a rich video embedding capable of capturing subtle changes in different frames of the video.


Given a video $V \in \mathbb{R}^{t . f_{sv} \times W \times H \times 3 }$ of duration $t$ seconds, recorded at $f_{sv}$ frames per second (fps), our goal is to generate a video embedding $E_V \in \mathbb{R} ^ {T \times h} $, where $T = t \times f_{sv}$ and $h$ is the size of feature vector for each frame $\textit{f}_i$ in the video. 
We use CLIP to capture semantically rich and meaningful information from each frame of the video. 
The image encoder converts each frame of the video into a feature vector $ E_V^i = \EuScript{F}_V ( f_i ) $. 
We stack all such feature vectors for all the frames in the video to get the final feature representation of the video $ E_V = \{ E_V^i \} _{i=1}^{N= t . f_{sv} } $. 
This feature representation of the video can be interpreted as an image of size $T \times h$ as shown in \Cref{fig:architecture}. 

Images sharing the same embedding space as text is crucial for our model design as explained in the further sections.
%
$
     E_V = \{ E_V^i \} _{i=1}^{N= t . f_{sv} } = \{ \EuScript{F}_V ( f_i ) \} _{i=1}^{N} = \{ \EuScript{F}_{V-CLIP} ( f_i ) \} _{i=1}^{N} .
$

We set the audio sampling rate and video frame rate such that they have the same $T$.




\begin{figure*}[!t]
\centering
\includegraphics[width=6.0in]{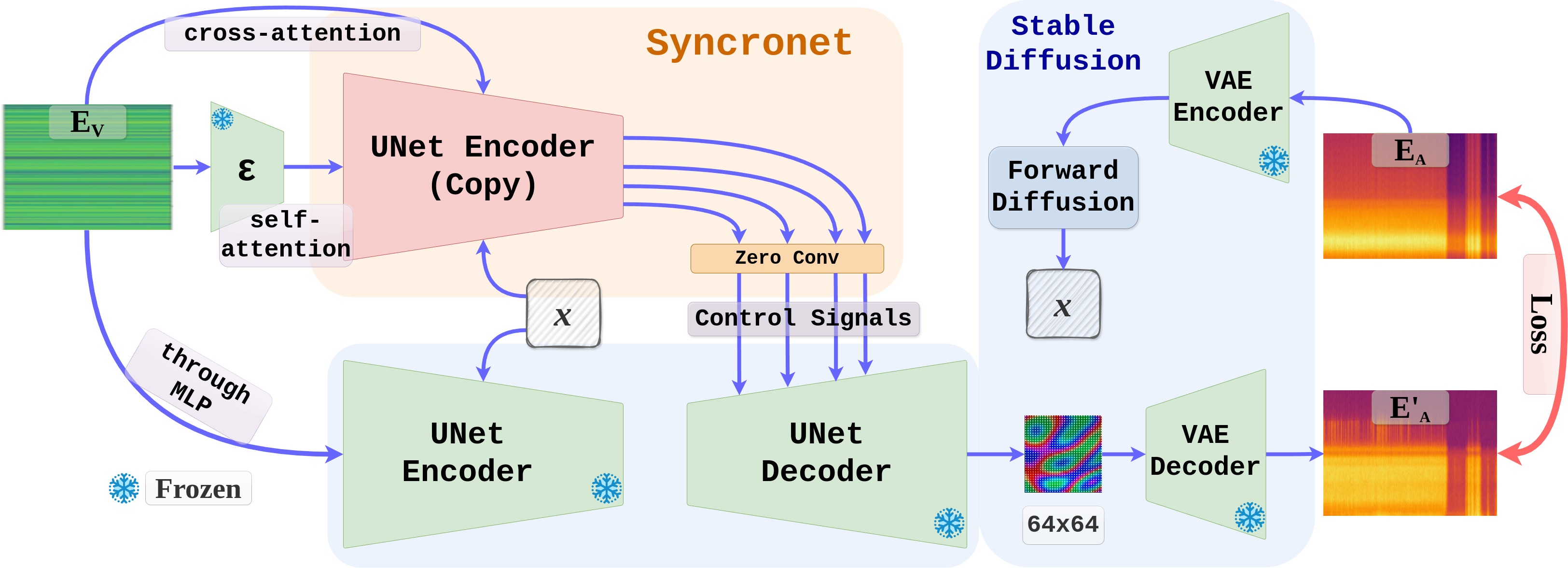} 
\caption{Figure describes the training of Syncronet model. A trainable copy of Stable Diffusion's UNet encoder generates control signals through zero convolution layers, providing pixel-level control to the pretrained UNet Decoder model. The UNet decoder generates a $64x64$ encoded feature map, which goes through VAE decoder to generate the predicted audio spectrogram $E'_A$.}
\label{fig:syncronet}
\vspace{-4mm}
\end{figure*}

\subsection{Diffusion Model for Audio Synthesis} 
\label{subsec:ldm}
Since we represent audio as spectrograms, we can build on the strengths of Latent Diffusion Models~\cite{stable_diffusion} for high-quality audio synthesis.
LDMs (e.g.,~Stable Diffusion) use pretrained Variational Autoencoder~\cite{transformet_vae} to convert the original image $x \in R^{H\times W\times 3}$ into a compact latent representation $z \in R^{h\times w\times c}$, where the forward and reverse diffusion processes are applied~\cite{stable_diffusion}. 
The decoder then converts the compact latent representation back to pixels. 


Stable Diffusion (SD) can use text prompt conditioning to guide the image generation. 
The text prompts are given to a pretrained CLIP text encoder to generate a text embedding $c_t \in \mathbb{R}^{77 \times 768} $, which is also given to the UNet encoder of SD. 
Thus, the overall objective function for SD becomes:
\begin{equation}
    \EuScript{L}_{LDM} := \mathbb{E}_{\epsilon(x),\epsilon \sim \EuScript{N}(0,1), t, c_t } [ || \epsilon - \epsilon_\theta ( z_t, t, c_t ) ||_2^2 ]
\end{equation}
where, $\epsilon_\theta (\cdot, t)$ is a time-conditional U-Net~\cite{unet} model, $\EuScript{N}(0,1)$ is the Normal distribution, $z_t$ is the latent code, and $c_t$ is the text embedding.
%

We found Stable Diffusion's VAE to be very effective in reconstruction and generating spectrograms. The same observation was made in \cite{diff_foley}.
In our model, we replace the text prompt from the SD with the video embedding $E_V$. 
The mean of all the image embeddings $\frac{1}{N} \sum_{i=1}^N E_V^i$ is passed through a small MLP to map it to the corresponding $c_t$ providing some contextual information to the UNet encoder blocks of the SD. 
As our video embeddings $E_V$ are also generated using CLIP, it makes it easier to reuse existing SD models that already operate with CLIP embeddings.
More details about the MLP are given in the supplementary.

To generate synchronized audio from video, we use an additional conditioning $C$ coming from our \textbf{\modulename} model. 
This additional conditioning provides the pixel-level control (i.e.,~\textbf{local guidance}) and acts as a local conditioning mechanism between $E_V$ and the SD's feature space.
Thus, the final SD model is conditioned on mean $E_V$ providing some global context to the UNet encoder, and $C$ for adding local pixel-level control to the UNet based decoder for controlled audio synthesis (see \Cref{fig:architecture}(left)).
Thus, the final overall objective function for the training becomes:
\begin{equation}
    \EuScript{L}_{LDM} := \mathbb{E}_{\epsilon(x),\epsilon \sim \EuScript{N}(0,1), t, E_V, C } [ || \epsilon - \epsilon_\theta ( z_t, t, E_V, C ) ||_2^2 ].
\end{equation}
Next section describes how we use $E_V$ for additional conditioning $C$ using the \textit{Syncronet} model.


\subsection{Time-Aware Audio Control using \textit{\textbf{\modulename}}} 
\label{subsec:syncronet}
We propose an strategy to learn the correspondences between the input video embeddings and the audio frequencies for every time step $t$ using \textbf{\modulename}.
\modulename builds upon ControlNet~\cite{controlnet}, a modal to enhance large pretrained text-to-image diffusion models with spatially localized, task specific image conditions for pixel-level control.
Originally, ControlNet was designed for a variety of spatial conditions like Canny edges, Hough lines, user scribbles, human key points, segmentation maps, shape normals, depths, and cartoon line drawings.
ControlNet 
works by processing these spatial conditioning and injecting additional control signals to the pretrained diffusion models.


\modulename enhances the original ControlNet architecture to replace the text embedding $c_t$ with the video embedding $E_V$, and uses self-attention and cross-attention mechanism to inject time-aware control signals to the Stable Diffusion's UNet Decoder (see \Cref{fig:syncronet}).
As a first step, we make a trainable copy of the entire UNet based encoder of SD along with the middle block and initialized them with the pretrained weights of SD 2.1.
The goal here is to use this trainable encoder to generate control signals that can be plugged into the pretrained SD's UNet decoder blocks providing pixel-level control to output (see Fig.~\ref{fig:syncronet}). The trainable copy is connected to the frozen SD through zero convolution layers to avoid any influence of noisy control signals during the start of the training.

\begin{wrapfigure}{r}{0.25\textwidth}
\vspace{-4mm}
\centering
\includegraphics[width=0.25\textwidth]{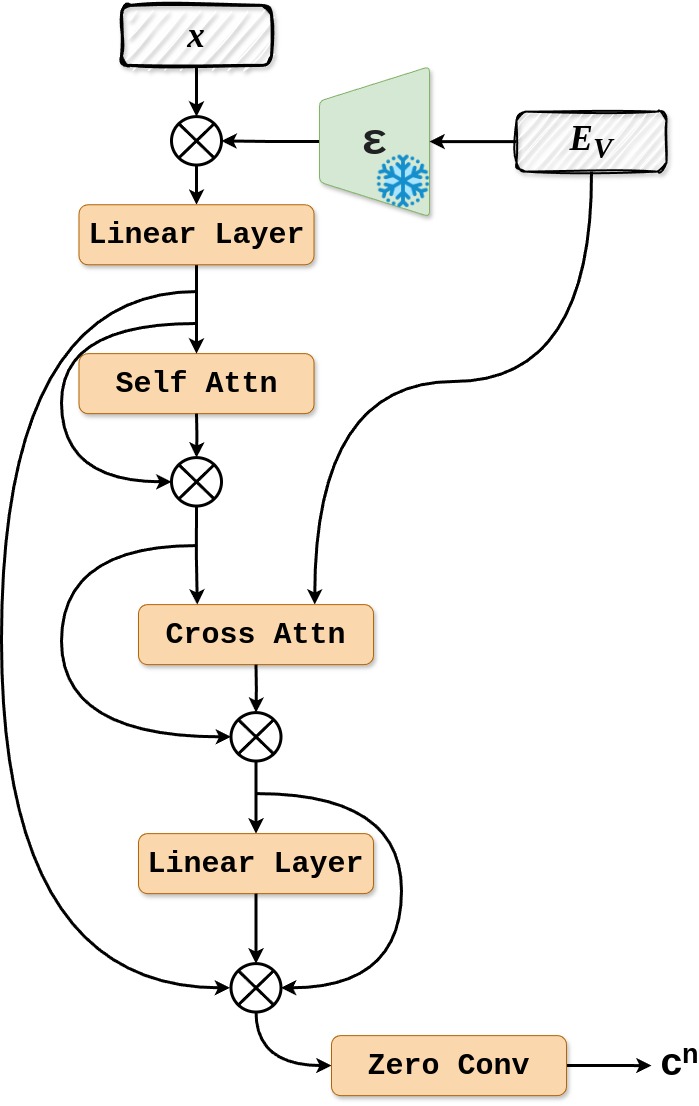}
\caption{\label{fig:cross_attn}Self-attention is applied to the sum of encoded video embedding and the noisy input sample. Then the cross-attention is applied between $E_V$ and the previous intermediate signal. Post which, they are passed through a linear layer followed by a zero convolution layer to get control signal $c^n$.
}
\end{wrapfigure}

As shown in Fig.~\ref{fig:cross_attn}, first, we transfer the input video embedding $E_V$ to a more rich $64 \times 64$ feature space through a pretrained encoder $\epsilon(.)$, where it's added to the input noisy data sample $x$. $h = x + \epsilon(E_V)$.
The added sum is further enriched by passing them through a self-attention block.
\vspace{-2mm}
\begin{equation}
     Self-Attn(Q_h, K_h, V_h) = Softmax ( \frac{Q_h K_h^T}{\sqrt{d_K}} ) V_h
\end{equation}
\vspace{-4mm}
\begin{center}
    $ h = h + Self-Attn(Q_h, K_h, V_h) $
\end{center}
where, $Q_h$, $K_h$, $V_h$ represents the Query, Key, and Value matrices derived from $h$. This would help incorporate contextual information from spatially varying signals in the same feature space, allowing the model to generate quite relevant audio. 

However, this alone is not sufficient to capture the fast temporal changes in the video.
Thus, alongside self-attention with the encoded video embedding $\epsilon (E_V)$, we also apply the cross-attention with the original video embedding $E_V$ to guide the audio spectrogram generation directly with the changes in time domain. 
This will help the model effectively learn the synchronization between the time domain and the rich feature space of Stable Diffusion. 
Also enabling the model avoid any inherited time-agnostic properties in the encoder model of SD, which has never seen time aware data.
\begin{equation}
     Cross-Attn(Q_h, K_V, V_V) = Softmax ( \frac{Q_h K_T^T}{\sqrt{d_K}} ) V_V
\end{equation}
\begin{center}
    $h = h + Cross-Attn(Q_h, K_V, V_V) $
\end{center}
where, $Q_h$, $K_V$, $V_V$ represents the Query, Key, and Value matrices derived from $h$ and $E_V$, respectively.
After passing through a \textit{linear layer}, a \textit{zero convolution} layer is applied to get the control signal $c^n$.
This attention mechanism (Fig.~\ref{fig:cross_attn}) is applied to all the Spatial Transformers in the Stable Diffusion's UNet encoder and middle blocks ($7$ in total). More details are given in the supplementary.

Similar to the original ControlNet model, the \modulename model $\EuScript{S}$ generates $13$ control signals using $x$ and $E_V$. $
     C := \{ c^n \}_{n=1}^{13} = \EuScript{S} ( x, \epsilon(E_V), E_V ).
$
These control signals are added to the $12$ skip-connections and $1$ middle block of the Stable Diffusion's UNet decoder block providing local pixel-level guidance.

\subsection{Training and Inference}


Before training \modulename, we train the video-to-text embedding MLP, and initialized SD 2.1, \modulename encoder~$\epsilon(.)$, and CLIP image encoder with their original pretrained weights. All these module are kept frozen during the \textit{\modulename} training.

For training \modulename first, the video frames are encoded using the video encoder. The video embedding is then given to the \textit{\modulename} which generates control signals to guide the SD's UNet decoder. 
The UNet decoder generates a $64 \times 64$ feature representation, which then goes to the VAE decoder to generate the audio spectrogram. The loss is applied between the generated audio spectrogram $E'_A$ and the ground truth spectrogram $E_A$ as shown in fig.~\ref{fig:syncronet}.
The Syncronet training is carried out out as per the objective mentioned in Eq.~5. 
Training has been done using the same loss functions . 
We use DDIM~\cite{ddim} for faster and consistent sampling. 
Upto 1000 time steps were used in the forward diffusion process, and 20 during the denoising. 
Training was done using AdamW optimizer with a learning rate of $1e-4$ for 50 epochs.
Once, the model is trained, we can sample new objects from the learned 

%
During \textbf{inference}, the given video sampled at 30 fps is given to the CLIP video encoder to obtain an image-like video embedding. 
Syncronet generates the control signals using this video embedding, which is then given to a DDIM sampler \cite{ddim} which predicts the conditioned encoded feature map for audio spectrogram.
This feature map is decoded using Stable Diffusion's pretrained decoder to obtain the audio spectrogram synchronized with the input video.
The pretrained audio super-resolution (ASR) model is used the enhance the generated audio quality and recover the losses occurred due to rescaling.
More details on ASR are given in the supplementary.
We scale and invert this enhanced audio spectrogram using Griffin-Lim~\cite{griffin_lim_faststft} algorithm to obtain the audio waveform of 10 seconds. 
The average infernce time for our method is at par with the current SOTA V2A model~\cite{diff_foley}. The inference time comparison is presented in Table~\ref{table:fid_compare}.

\vspace{-3mm}

\paragraph{Video-Audio Alignment Score (VAAS)}
Inspired by \cite{diff_foley}, we propose
a Video-Audio Alignment Score (VAAS) to compare the audio-visual alignment. 
As there is no standard benchmark to measure the alignment accuracy, we trained a classifier that uses ViT to extract the audio and video features and calculates the alignment between them. 
The classifier training dataset consists of three AV pairs - 50\%, 25\%, and 25\%, respectively for pairs from the same video, pairs from different videos, and the pairs of the same video but temporally shifted by a random amount. 
Only the first 50\% AV pair are considered as $True$ during the training and the remaining 50\% are considered $False$. 
The ViTs for extracting AV features share the same weights.



\begin{figure*}[!ht]
\centering
\includegraphics[width=5.8in]{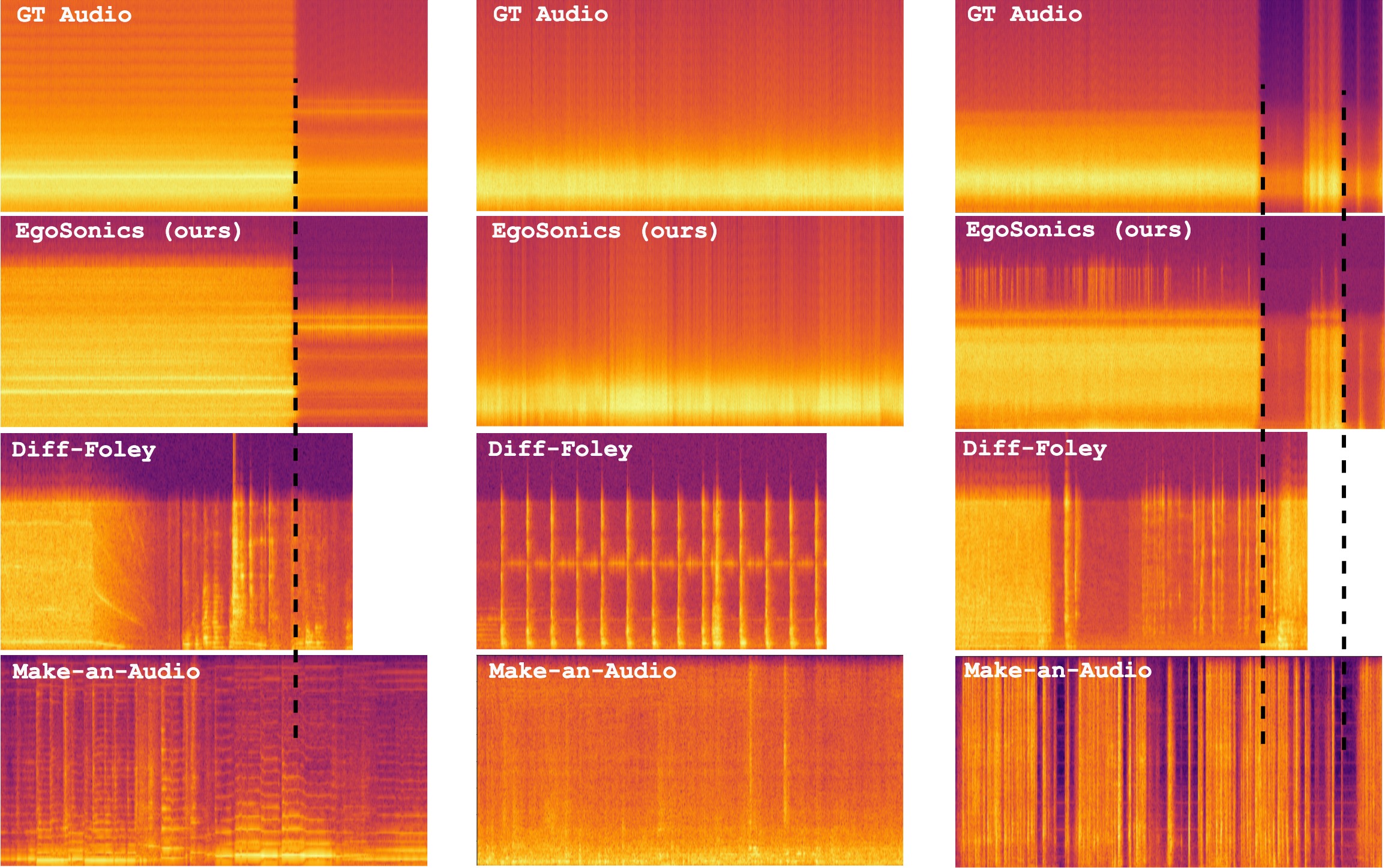} 
\caption{Figure compares how well each model can synchronize the generated audio with the GT audio. Our method can generate very synchronized audios with higher quality. On the other hand, Diff-Foley and Make-an-Audio fails to do so, and often fails to synthesize contextually correct audio. Diff-Foley can only synthesize 8 seconds long audio compared to 10 seconds for others.
}
\label{fig:sync}
\end{figure*}

\section{Experiments}
\label{others}

\vspace{-2mm}

The goal of our experiments is to compare the audio and synchronization quality of generated audio, justify and ablate key design decisions, and demonstrate the application of our method in improving video summarization.
\vspace{-3mm}
\paragraph{Dataset.} 
We use one of the largest audio-visual egocentric datasets, Ego4D~\cite{ego4d}, which consists of roughly 3600 hours of egocentric videos capturing various daily activities.
However, not all videos in this dataset contain audio, and even those with audio do not always contain useful learnable information (e.g.,~background noise).
Therefore, we focus on videos with useful learnable visual audio cues, and pick 10-second chunks as described in \Cref{subsec:avprep}.
After this process, we obtain around 150K videos with the corresponding audio.
We use STFT~\cite{perraudin2013fast} to convert these 10 seconds long audio waveform into spectrograms.
We sampled the audio at 22~KHz to get $E_A \in \mathbb{R} ^ {430 \times 1024}$.

\vspace{-2mm}
\paragraph{Experimental Setup.} 

\begin{table*}[h]
\centering
\caption{\textit{V2A generation results comparing various methods over 3 metrics (FID, IS, VAAS). We have also compared them against the audio sampling rate (ASR), maximum frequency support (MaxF), spectrogram size used (Size), the length of generated audio (AL) in seconds, the number of video frames used for guidance (VF), and inference time (IT) in seconds. Our method, using 30 fps, sampled at highest frequency can generate the highest resolution spectrogram for a 10 seconds long audio. It also outperforms the existing V2A on FID, IS, and VAAS.}}
\scalebox{0.9}{\begin{tabular}{c|c|c|c|c|c|c|c|c|c}
 \textbf{Method} &\textbf{ ASR} & \textbf{MaxF} & \textbf{Size} & \textbf{AL (sec)} & \textbf{VF} & \textbf{FID} $\downarrow$ & \textbf{IS} $\uparrow$ & \textbf{VAAS} $\uparrow$ & {\textbf{IT (sec)}} \\
 \hline
 \hline
 Im2Wav~\cite{sheffer2023hear} & $16 KHz$ & $8 KHz$ & $-$ & $5$ & $1-30fps$ & $91.73$ & $19.15$ & $59.24$ & {$22$} \\
 Make-an-Audio~\cite{make_an_audio} & $16 KHz$& $8 KHz$ & $80 \times 624$ & $9-10$ & $1-4$ & $83.72$  &  $22.14$  & $64.57$ & {$17$} \\
 Diff-Foley~\cite{diff_foley} & $16 KHz$ & $8 KHz$ & $256 \times 128$ & $8$ & $4fps$ & $76.92$ & $37.92$ & $78.19$ & {$1.9$} \\
 \shortname (ours) & $22 KHz$ & $20 KHz$ & $512 \times 512$ & $10$ & $30fps$ & $\textbf{41.14}$ & $\textbf{54.41}$ & $\textbf{92.77}$ & {$2.6$} 
\label{table:fid_compare} 
\end{tabular}}
\vspace{-4mm}
\end{table*}
We train our model for 50 epochs on 8 Nvidia 3090 GPUs. 
Video embeddings and audio spectrograms are extracted from the 10-second clips and resized to 512$\times$ 512 to ensure synchronization between video and audio modalities.
We used the pretrained weights of CLIP Image encoder and used it to obtain $E_V$.
\modulename is initialized with the pretrained SD's UNet encoder weights, and the zero convolutions are initialized with zero weights to prevent issues in the generated results during initial iterations. 
We also use the same Classifier-Free Guidance Resolution Weighting scheme as proposed in \cite{controlnet}.

\begin{table}[ht]
\centering
\caption{\textit{Comparing the effect of using audio generated from different methods on improving video summarization task. It is interesting to note that, even the unsynchronized audios generated from our model have sufficient contextual information to outperform the existing methods.
}}
\vspace{-2mm}
\scalebox{0.9}{\begin{tabular}{c|c}
 \textbf{Method} & \textbf{Cosine Similarity} $\uparrow$ \\
 \hline
 \hline
 Without Audio &  $0.67$ \\
 GT Audio &  $0.81$  \\
 \hline
 Im2Wav~\cite{sheffer2023hear} &  $0.57$  \\
 Make-an-Audio~\cite{make_an_audio} & $0.68$ \\
 Diff-Foley~\cite{diff_foley} &  $0.71$  \\
 EgoSonics w/o Cross-Attn &  $0.72$ \\ 
 EgoSonics (ours) &  $\textbf{0.76}$
\label{table:vid_sum} 
\end{tabular}}
\vspace{-6mm}
\end{table}
\vspace{-3mm}
\paragraph{Metrics.}
For quantitative evaluation, we have used the two standard Frechet Distance (FID) and Inception Score (IS) as used in most previous works \cite{diff_foley, make_an_audio, sheffer2023hear}. FID is used to measure the distribution-level similarity with the dataset, IS is effective is measuring the sample quality and diversity. The proposed VAAS metrics is essential to measure how well the generated audio is synchronized with the video. Thus, it captures the alignment between the audio and video features.

\subsection{Qualitative and Quantitative Comparisons} 
Qualitative results generated from our pipeline are shown in \Cref{fig:results_1} (please also see supplementary video).
From the results, we see that our method not only generates good quality audio spectrograms corresponding to video activities but also achieves synchronization as depicted by the dotted lines where some distinct event in the video.
Moreover, we observe that our method implicitly learns to suppress background noise and only generate sounds of interest, thus acting as an \textbf{Implicit Noise Removal} technique.
This is the main reason why all the ground truth (GT) audios contain a lot of soft frequency components, but audio generated from our method does not. 
More results are presented in the supplementary.

The quantitative results of our method are presented in Table~\ref{table:fid_compare}.
Our proposed method generating 10 seconds long audio with frequencies upto 20~KHz is able to generate better results than all the baselines in FID and IS measurement. Using 30 frames per second from the input video helped us in better alignment of the generated audio, thus, achieving a 14$\%$ increase in Video-Audio Alignment Score (VAAS).

\subsection{Video Summarization} 
To demonstrate applications of our method beyond just audio generation, we show that it can be used to improve the quality of video summarization. 
Video summarization \cite{video_summ} 
involves providing a short summary of a scene in a given video via understanding the video frames. 
We leverage the fact that audio provides additional cues about the scene~\cite{audio_video_floor1, audio_video_floor2} through the easily identifiable sounds associated with each activity.
Our aim is to incorporate the audio generated from our method alongside video encoding, and observe an improvement in the accuracy of video summarization. 
This will highlight the effectiveness of our method in generating relevant sounds capturing high-order details of the scene.
More details about the video summarization architecture are presented in the supplementary.
Table~\ref{table:vid_sum} compares the cosine similarity between the text embedding predicted from the video summarization model and the GT text embedding we generated using Ego4D narrations \cite{ego4d} for various methods.
It can be seen that the audio generated by EgoSonics can significantly help in predicting the correct text embedding and outperforms existing V2A methods.

\begin{table}[h]
\vspace{-2mm}
\centering

\caption{\textit{Comparing the performance of our model against various changes, including removing the \textit{Syncronet} model, removing cross-attention in \textit{Syncronet}, and using video guidance at different fps.}
}

\scalebox{0.9}{\begin{tabular}{c|c|c}
 \textbf{Method} & \textbf{FID} & \textbf{VAAS} \\
 \hline
 \hline
 EgoSonics w/o Syncronet &  $150.28$  & $-$ \\
 EgoSonics w/o Cross-Attn & $\textbf{34.33}$ & $72.83$ \\
 EgoSonics $@4$ fps & $37.18$ & $84.97$ \\
 EgoSonics $@15$ fps & $40.72$ & $89.39$ \\
 EgoSonics $@30$ fps & $41.14$ & $\textbf{92.77}$

\label{table:ablation} 
\end{tabular}}
\vspace{-4mm}
\end{table}

\subsection{Ablation Study}

\paragraph{w/o \modulename.}
Our method achieves better audio-visual synchronization since we use 30 frames per second to make \modulename learn the per-frame correspondences between audio and video. 
Here, we analyzed the effect of removing \modulename completely and solely using the mean video embedding to guide the Stable Diffusion. 
From the results, it can be seen that there is a significant drop in FID meaning the pre-trained Stable Diffusion cannot synthesize the good quality audio spectrograms. 
Consequently, SD is not able to generate conditional audio and lacks synchronization.

\vspace{-2mm}
\paragraph{w/o Cross-Attention in \modulename.}
Now, we also analyzed the effect of removing Cross-Attention with the video embedding $E_V$ in \modulename and just using Self-Attention as given in Eqn.~3.
From \Cref{table:ablation}, it's clear that our method is able to synthesize good quality audio spectrograms as the FID score is good. However, the poor alignment score indicates that the local pixel level alignment control to achieve synchronization is coming from the control signals through cross-attention with the video embedding $E_V$ in the time domain.

\vspace{-2mm}
\paragraph{Effect of Video Rate on Alignment.}
Unlike previous methods using an average video embedding, or 1-4~fps for audio generation, we use 30~fps to extract the temporal information from the video.
In this section, we analyze the impact of using 30 frames on AV synchronization.
We compare the performance of \modulename at different fps including 4~fps used by the current SOTA Diff-Foley \cite{diff_foley}.
From \Cref{table:ablation}, it can be seen that our method not only generates good quality audio spectrograms, but also achieves a VAAS score of 85.
As we increase the fps to 15, the alignment score increases to 89, and then to 92.77 with 30~fps.



\vspace{-2mm}
\subsection{User Study}
\vspace{-2mm}
We conducted an user study for a subjective evaluation of our synthesized audios. With randomly selected 15 participants from different backgrounds and asked them to do a 10 minutes survey. The survey present 16 stimulus to the participants and ask them to rate their realism using a predefined scale of 1-5. Participants were also asked if the audio is relevant to the video, it's synchronization, and if it seems to be from a different category (e.g., a carpentry sound coming from a vacuum cleaner). The following are the outcomes of this study:
80$\%$ of the users believe that 90$\%$ of our audios are realistic with an Mean Opinion Score (MOS)~\cite{MOS} of more than $4.0$.
In 100$\%$ of the cases, users preferred our audio over the current SOTA V2A model (Diff-Foley).
In more than 75$\%$ cases, users could not distinguish our audio from the GT audio, and rated both to be realistic.
In more than $80\%$ of the cases, users did not find our audios to be belonging to another category. More in Supplementary A.

\vspace{-2mm}

\section{Conclusion}
\vspace{-2mm}
In this paper, we propose a novel audio-visual framework called EgoSonics. Much like how humans effortlessly connect sound with visual stimuli, EgoSonics has the ability to predict synchronized audio for muted Egocentric videos. Leveraging a streamlined architecture that incorporates generation capabilities of Stable Diffusion, EgoSonics holds promise for various applications. The proposed \textit{Syncronet} can generate control signals from input video frames to provide pixel-level control to Stable Diffusion. Furthermore, we showcased our method's superior performance over existing methods and illustrate how this approach can enhance downstream applications like video summarization.

\vspace{-4mm}
\paragraph{Limitations and Future Work.}

Although, EgoSonics can generate good quality audio for egocentric videos, there are a few limitations. One thing we observed in some of our generated samples is the misalignment due to the lack of rich visual information resulting from occlusions. Our approach is also limited by the amount of training data. For example, there are a very few samples of musical instruments in the Ego4D dataset, consequently, our model doesn't perform very well on such videos. We believe that such challenges can be solved by training our model with millions of audio-video pairs.

{
    \bibliographystyle{ieeenat_fullname}
    \bibliography{main}
}

\clearpage

\section*{EGOSONICS - SUPPLEMENTARY}
\label{supplementary}

\section*{A. User Study}
\label{supp_user_study}

We conducted an user study for a subjective evaluation of our synthesized audios. With randomly selected 15 participants from different backgrounds and asked them to do a 10 minutes survey using Google Forms. The survey presented 16 videos with audios to the participants and asked them to rate their realism on a scale of 1-5 (1 being "Doesn't Sound Real" and 2 being "Sounds Real"). Participants were also asked if the audio is contextually relevant to the video, if the events in the video and the corresponding audio are synchronized. To verify that our method doesn't overfit to one category of sounds, we also asked the participants to evaluate if the audio seems to be from a different category (e.g., a carpentry sound coming from a vacuum cleaner).

The following are the outcomes of this study:

\begin{enumerate}
    \item 80$\%$ of the users believe that 90$\%$ of our audios are realistic with an Mean Opinion Score (MOS)~\cite{MOS} of more than $4.0$. This means that EgoSonics is able to generate realistic audios.
    \item In 100$\%$ of the cases, users preferred our audio over the current SOTA V2A model (Diff-Foley). This means all the 15 users believe that all of our audios are better that the baseline.
    \item In more than 75$\%$ cases, users were unable to distinguish our audio from the GT audio, and rated both to be realistic. This again proves that EgoSonics is able to generate realistic and synchronized audio.
    \item In more than $80\%$ of the cases, users did not find our audios to be belonging to another category, meaning that EgoSonics doesn't overfit to any particular class and is able to generate all kinds of daily activity audios.
\end{enumerate}

\begin{figure*}[!ht]
\centering
\includegraphics[width=5.5in]{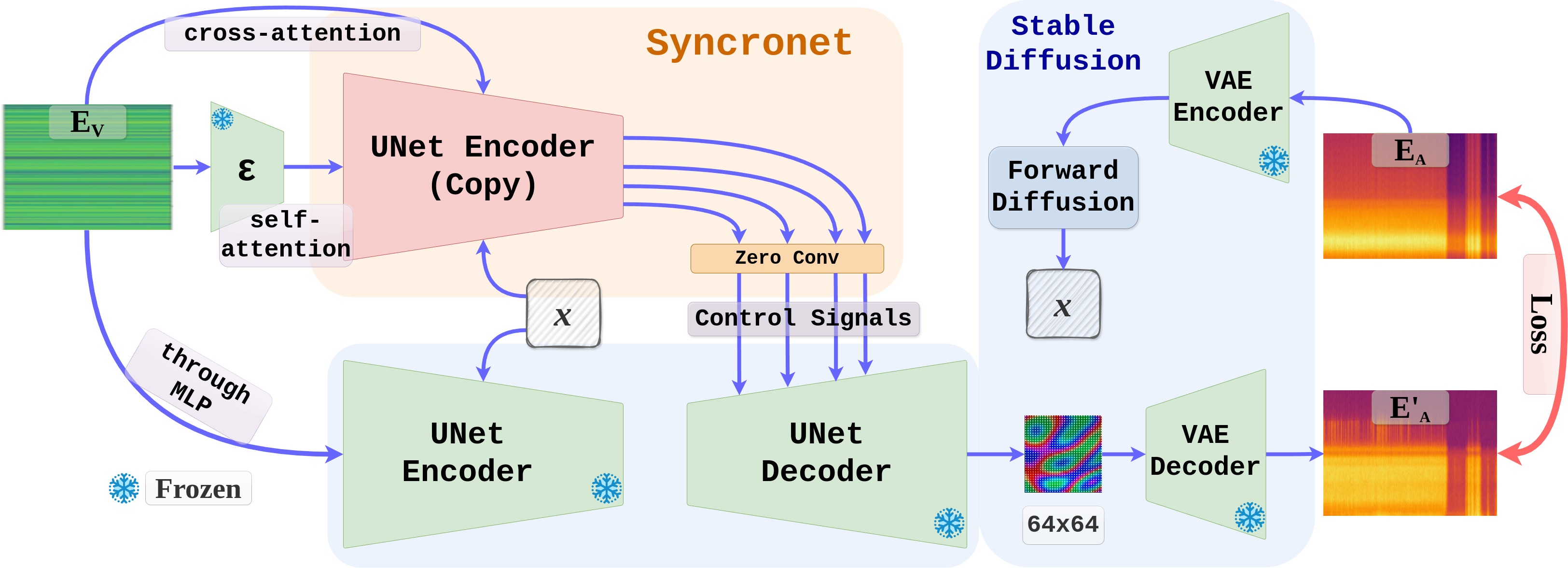} 
\caption{\textit{Syncronet} training.}
\label{fig:architecture_supp}
\end{figure*}

\section*{B. Syncronet}

We propose \textit{Syncronet}, a model to learn the correspondence between audio and video modalities by learning correlation between the input video embeddings and the audio frequencies for every time step $t$. 
ControlNet~\cite{controlnet} is a current state-of-the-art neural network architecture that was introduced to enhance large pretrained text-to-image diffusion models with spatially localized, task specific image conditions providing pixel-level control. 
Over the time, ControlNet has been shown to work for a variety of controlled image generation tasks including but not limited to spatial conditions like \textit{Canny edges, Hough lines, user scribbles, human key points, segmentation maps, shape normals, depths, cartoon line drawings}, etc. ControlNet basically works by processing these spatial conditioning and injecting additional control signals to the pretrained diffusion models. 
However, generating time-aware control signals to guide the output in temporally consistent manner has not been extensively studied using ControlNet. In fact, to the best of our knowledge, only Music ControlNet~\cite{music_controlnet} uses it to generate partially-specified time-varying control signals. In this paper, we modified the ControlNet architecture to operate over given video embedding in both encoded feature space, and time space to generate control signals that can provide local pixel-level control to the Stable Diffusion's UNet model to generate time-consistent audio spectrograms $E_A$. The generated spectrograms possess a strong correlation with the input video embedding and this results in highly synchronized audio of daily activity videos - where most previous methods fail due to design constraints.

We use \textit{Syncronet} to provide control signals to Stable Diffusion (SD) $2.1$. As a first step, we make a trainable copy of the entire UNet based encoder of SD along with the middle block and initialized them with the same pretrained weights of SD 2.1. The goal here is to use these trainable encoder to generate control signals that can be plugged into the pretrained SD's UNet decoder blocks providing pixel-level control to output (see Fig.~\ref{fig:architecture_supp}). The trainable copy is connected to the frozen SD through zero convolution layers to avoid any influence of noisy control signals during the start of the training.

Similar to ControlNet, the input conditioning image ($E_V$) of size $512 \times 512$ is converted to a feature space of size $64 \times 64$, that matches the feature space of Stable Diffusion, through a pre-trained image encoder $\epsilon$. We used the same convolution based image encoder as used in \cite{controlnet}, and initialized it with the same weights. The image encoder is kept frozen throughout the training. 
The encoder $\epsilon$ extracts a feature space vector $c_f$ from the input conditioning $E_V$. 

As shown in Fig.~\ref{fig:architecture_supp}, the noisy sample $x$ is generated through forward diffusion process, where the input audio spectrogram of size $512 \times 512$ is encoded to a feature vector of size $64 \times 64$ through pre-trained VAE encoder. We used the same pretrained VAE encoder-decoder network as SD and initialized them with the same weights. Noise is added to the feature vector to get a noisy sample $x$ for $T=1000$ steps. 
Then the encoded video embedding $\epsilon(E_V)$ is added to the input noisy data sample $x$. $h = x + \epsilon(E_V)$.
The added sum is further enriched by passing it through a self-attention block.


\begin{equation}
     Self-Attn(Q_h, K_h, V_h) = Softmax ( \frac{Q_h K_h^T}{\sqrt{d_K}} ) V_h
\end{equation}

\begin{center}
    $ h = h + Self-Attn(Q_h, K_h, V_h) $
\end{center}

where, $Q_h$, $K_h$, $V_h$ represents the Query, Key, and Value matrices derived from $h$.
As we have seen in Table~\ref{table:ablation}, only using this self attention block alone is sufficient to generate good quality audio spectrograms with an FID of $34.33$. However, as the alignment score suggests, this alone is not sufficient to guide the diffusion model generate temporally consistent synchronized audio.

\begin{algorithm}
\caption{Generate Control Signals}
\label{algorithm1}
\begin{algorithmic}[1]
\REQUIRE $x$: Input Noisy Sample
\REQUIRE $EV$: Video Embedding
\REQUIRE $timesteps$: Timestep tensor
\ENSURE $control\_signals$: List of control signals

\STATE $t\_emb \gets \text{timestep\_embedding}(timesteps)$
\STATE $emb \gets \text{time\_embed}(t\_emb)$
\STATE $context \gets EV$

\STATE $guided\_hint \gets \text{encoder}(EV)$

\STATE $control\_signals \gets [\,]$

\STATE $h \gets x$

\FOR{$(module, zero\_conv) \in (\text{UNet.encoder\_blocks}, \text{Syncronet.zero\_convs})$}
    \IF{$module == \text{UNet.encoder\_blocks.first\_block}$}
        \STATE $h \gets \text{module}(h)$
        \STATE $h \gets h + guided\_hint$
        \STATE $guided\_hint \gets \textbf{None}$
    \ELSE
        \IF{$module$ is $\text{TimeStepBlock}$}
            \STATE $h \gets \text{module}(h, emb)$
        \ELSIF{$module$ is $\text{SpatialTransformer}$}  
            \STATE \COMMENT{\textcolor{blue}{Apply self-attention and cross-attention}}
            \STATE $h \gets \text{module}(h, context)$
        \ENDIF
    \ENDIF
    \STATE \text{Append} $\text{zero\_conv}(h, emb, context)$ \text{to} $control\_signals$
\ENDFOR

\STATE $h \gets \text{UNet.middle\_block}(h, emb, context)$
\STATE \text{Append} $\text{zero\_conv}(\text{UNet.middle\_block\_out}(h, emb, context))$ \text{to} $control\_signals$

\RETURN $control\_signals$
\end{algorithmic}
\end{algorithm}

Thus, to inject the temporal consistency to the control signals of \textit{Syncronet}, we also apply the cross-attention between the original video embedding $E_V$ and $h$ to guide the audio spectrogram generation directly with the time steps of $E_V$. 
This helps the model effectively learn the synchronization between the time domain and the rich feature space of Stable Diffusion, as we can see through a significant improvement in the alignment score. 
There are two options to apply cross-attention, either by using Query from $E_V$ and Key and Value matrices from $h$, or vice-versa. In our case, we use the latter approach as follows:

\begin{equation}
     Cross-Attn(Q_h, K_V, V_V) = Softmax ( \frac{Q_h K_T^T}{\sqrt{d_K}} ) V_V
\end{equation}

\begin{center}
    $h = h + Cross-Attn(Q_h, K_V, V_V) $
\end{center}

where, $Q_h$, $K_V$, $V_V$ represents the Query, Key, and Value matrices derived from $h$ and $E_V$, respectively.
After passing through a \textit{linear layer}, a \textit{zero convolution} layer is applied to get the control signal $c^n$.

The Stable Diffusion's UNet architecture contains 12 encoder, 12 decoder and 1 middle blocks. Similar to \cite{controlnet}, our trainable copy contains 12 encoder and 1 middle block consisting of several Vision Transformers (ViTs). Self-attention and cross-attention is applied in all the Spatial Transformers of encoder and middle blocks as described in algorithm~\ref{algorithm1}.

These control signals are added to the $12$ skip-connections and $1$ middle block of the Stable Diffusion's UNet decoder block providing local pixel-level guidance at $64 \times 64, 32 \times 32, 16 \times 16, 8 \times 8$ resolutions.

\section*{C. Dataset and Training}

Ego4D~\cite{ego4d} is a large scale multimodal dataset consisting of around $3600$ hours of daily activity videos. However, not every video in the dataset comes with corresponding audio, due to privacy concerns or technical limitations. Only half of the dataset has the corresponding audio. Now, when we look at the dataset, it contains a large amount of person-to-person conversations in shops, homes, outdoor scenes, etc., which doesn't serve any purpose in our use case. Thus, we only selected those videos that belongs to certain categories like \textit{cooking, carpentry, laundry, cleaning, working, farmer, mechanic, yardwork, blacksmith, etc.} We believe, only such categories are useful in learning corresponding between audio and video for day-to-day activities. Even, in these videos, not every section of the video is important as there's a lot of redundancy in the data. To overcome this limitation, we calculate the Root Mean Square (RMS) value of a 10 second clip randomly picked from the dataset as follows:

\begin{center}

    $S_{RMS} = \sqrt{ \frac{1}{N} \sum_{i=1}^N x_i^2 }$
    
\end{center}

where, $N$ is the total number of samples in audio waveform, and $x$ is the value of each sample. After calculating the RMS value of each $10$ second long audio sample, we compared it against a manually set threshold. If the sample's RMS value exceeded the threshold, we used it for training. This gave us a rich set of audio-video pair containing daily activities. We randomly picked a non-overlapping set of $150K$ such samples, which were used for training. Each $10$ seconds long video is sampled at $30$ frames per second and contains $300$ frames in total. The corresponding audio is sampled at $22KHz$, and converted to audio spectrogram using Short-Time Fourier Transform \cite{griffin_lim_faststft}. The audio spectrogram is resized to $512 \times 512$ from $430 \times 1024$ to make it compatible with Stable Diffusion's encoder-decoder network.

To get a more useful and compact video representation, we use video embedding as the representation of videos. Video embedding is a feature rich image-like representation of the video where each video frame is represented as a vertical vector of shape $1 \times 512$. $300$ such vectors are placed one after the other in the same sequence as frame number to get the video embedding of shape $300 \times 512$. Each video embedding is resized to $512 \times 512$ using bicubic interpolation to make sure it aligns with each time-step in the audio spectrogram.

Training of \textit{Syncronet} has been done using the same loss functions as ControlNet~\cite{controlnet}. We use DDIM~\cite{ddim} for faster and consistent sampling. Upto $1000$ time steps were used in the forward process, and 20 during the denoising. Training was done using AdamW optimizer with a learning rate of $1e-4$.

\section*{E. Audio Super-Resolution}

To upsample the generated audio spectrograms $E_A$ from a resolution of $512 \times 512$ to $512 \times 1024$, we trained s small 5 layer Convolutional Neural Network (CNN) with $16, 32, 64, 64, 32, 1$ output filters. Each layer is followed by ReLu activation and Batch Normalization. Sigmoid is used as the final activation to keep the values in $[0,1]$. The model was trained using pairs of original audio spectrograms of shape $512 \times 1024$, and their down sampled version of shape $512 \times 512$ using Mean Square Error (MSE) loss. Adam optimizer was used for the optimization and a learning rate of $0.001$ was used. The model was trained for $10$ epochs with a dataset of $200K$ audio samples.

\section*{F. Video Summarization}

\begin{figure*}[!ht]
\centering
\includegraphics[width=5.5in]{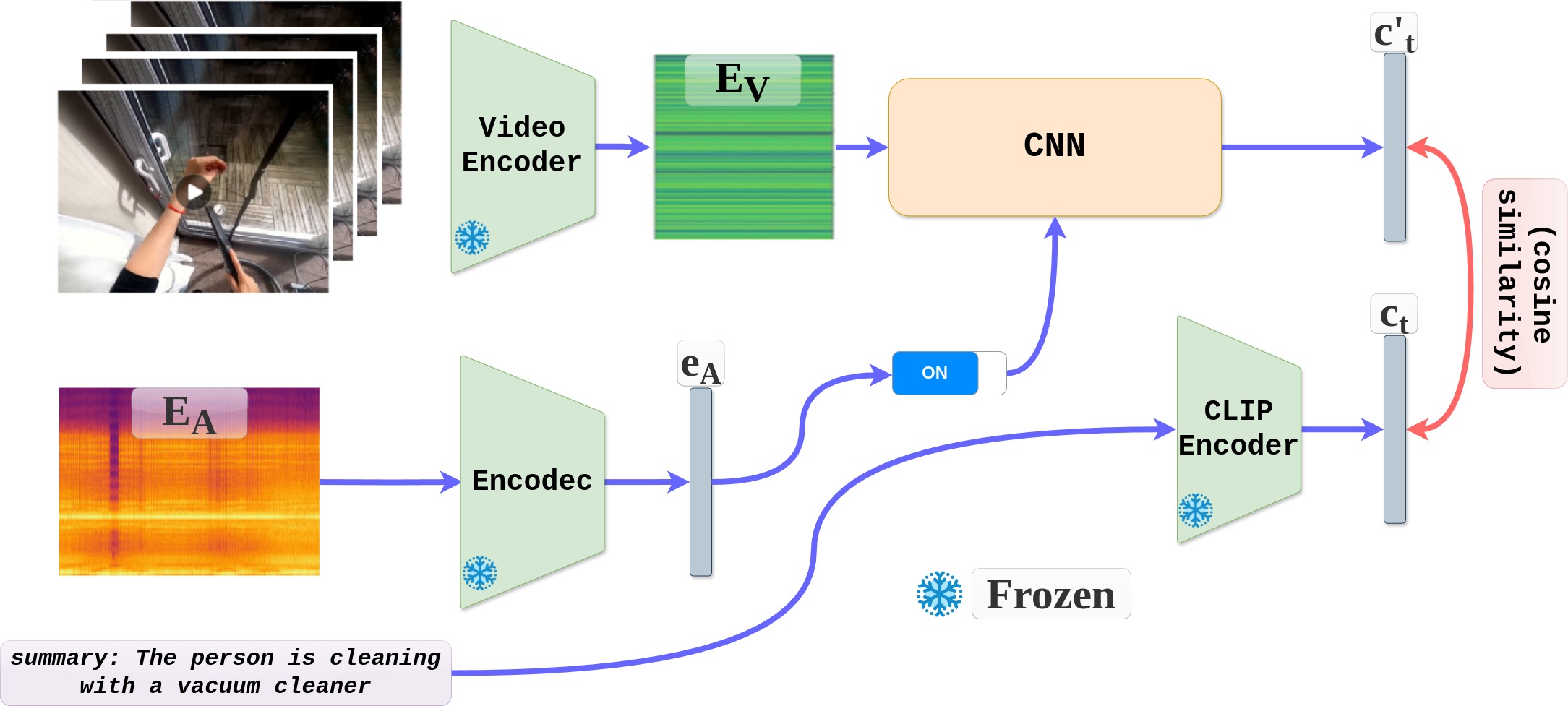} 
\caption{Video Summarization.}
\label{fig:supp_video_summ}
\end{figure*}

Video Summarization is a long studied task which involves providing a short summary of a scene in a given video. 
It has been widely accepted that the audio contains rich information about the scenes and can even provide sufficient cues to reconstruct a scene geometry~\cite{audio_video_floor1, audio_video_floor2}. 
We leverage this fact that audio provides additional cues about the scene through the easily identifiable sounds associated to improve the video summarization task.
Our aim is to incorporate the audio generated from our method alongside the input video frames, and observe an improvement in the prediction accuracy of corresponding scene summary.
We used a very simple setup to test this hypothesis as presented in Fig.~\ref{fig:supp_video_summ}. 
The input video frames are encoded to a rich feature space using the pretrained CLIP encoder.
Initially, when the toggle switch is in "OFF" state, the Convolutional Neural Network (CNN) takes the video embedding $E_V$ as the input, along with a zero vector as audio embedding $e_A$, and predicts the text embedding $c'_t$. 
The ground truth text embedding is estimated using the one sentence text summary after passing it through the CLIP text encoder. 
Ego4D dataset provides short narrations describing the activity of the scene. We used these narrations as the scene summary.
The CNN is trained using MSE loss and the parameters are optimized using Adam optimizer with a learning rate of $5e^{-3}$.

When the toggle switch is "ON", that is, when the the zero vector $e_A$ is replaced with the GT audio embedding, the CNN takes in this additional input and concatenates it with the video embedding vector through convolution. 
Similar as before, the model tried to predict the text embedding and the CNN is trained until convergence. 
We use a well known audio compression method EnCodec~\cite{encodec} to encode the audio waveform into a more rich neural codec representation.

Once the model is trained, we compared it's performance on the test dataset using various methods. The results are presented in Table~\ref{table:vid_sum}.

\section*{G. Video-to-Text Embedding MLP}

We trained a small two layer MLP that takes a normalized video embedding $E_V$, and generates a vector of shape $512$, which acts as a text embedding $c_t$ to the stable diffusion model. The MLP was trained using $200K$ pair of video and text embedding using MSE loss and Adam optimizer.

\begin{figure*}[!t]
\centering
\includegraphics[width=5.5in]{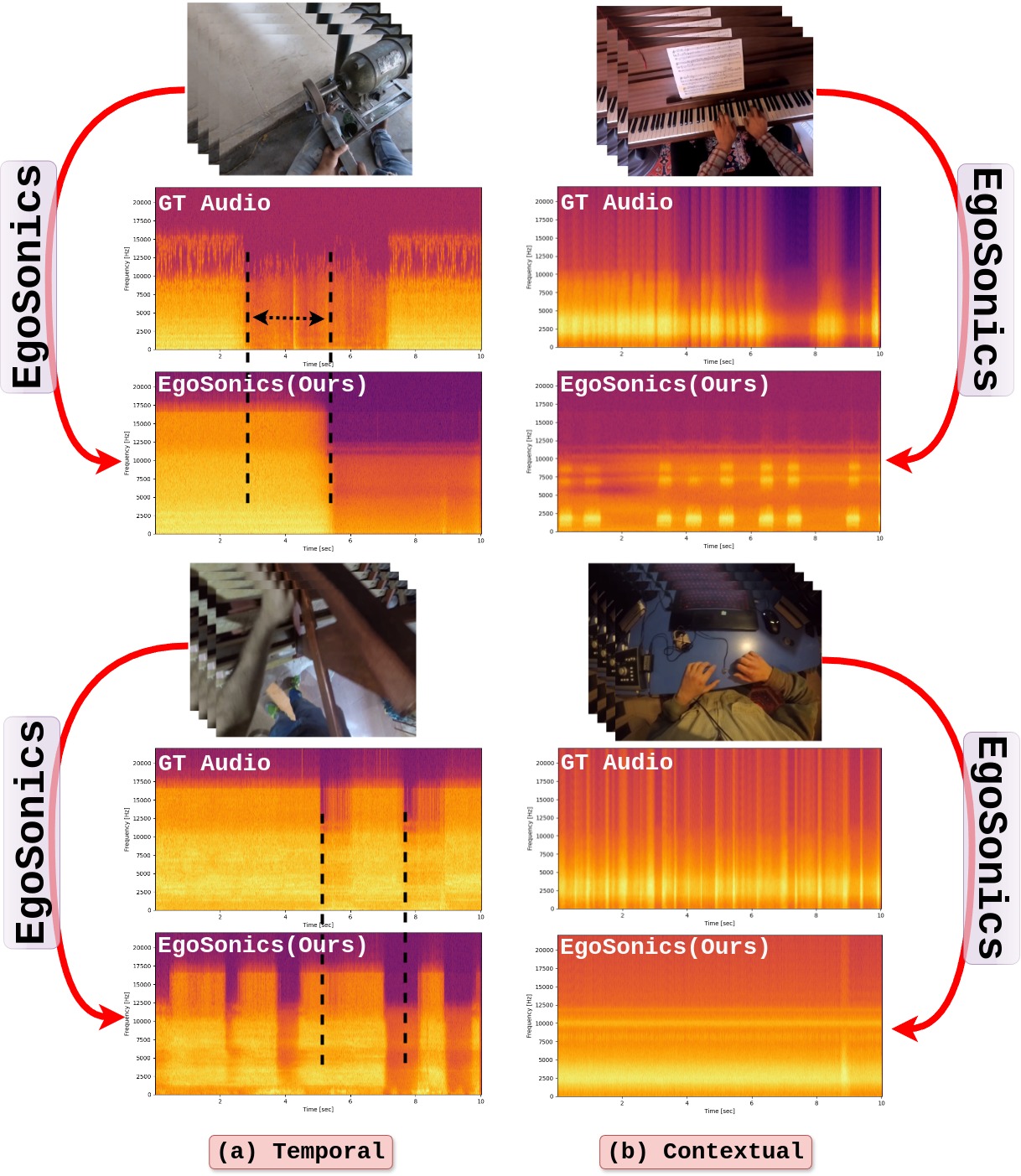} 
\caption{Failure Cases. There are two types of failure cases: (a) Temporal misalignment, (b) Contextual misalignment.}
\label{fig:supp_failure}
\end{figure*}

\section*{H. Synchronization Metrics}

An effective way of measuring the audio-video alignment is missing in the field of audio-visual learning. Thus, inspired by Diff-Foley~\cite{diff_foley}, we introduced a Vision Transformer based metrics (Alignment Score) that can calculate the synchronization between audio and video. Unlike Diff-Foley, we used ViT-B32 as a feature extractor to get audio features from $E_A$, and video features from $E_V$, and then use 5 Linear layers, each followed by a ReLu activation. We uses a pre-trained ViT-B32 trained on $IMAGENET1K\_V1$. The linear layers were trained using MSE loss and Adam optimizer with a learning rate of $0.0001$. Our training dataset consists of $200K$ samples, out of which $100K$ were labelled as $1$ and the remaining $100K$ as $0$. Audio samples belonging to the same video were all labeled as $1$ and accounts for $50\%$ of the training data. $25\%$ data is the audio samples randomly assigned with any video other than the original video. These were labeled as $0$. Remaining $25\%$ samples came from randomly shifting the audio anywhere between $1-5$ seconds from the true audio-video pair. These were also labeled as $0$. Our classifier reached an accuracy of $97\%$ on testing dataset comprising of $20\%$ of the training data kept separately. If the trained model classifies an audio-video pair as anything close to $1$, it means the two modalities are accurately aligned. \textit{EgoSonics} scores an average accuracy of $92\%$ on test dataset meaning that we significantly outperforms the existing methods in better synchronizing audio. 

For calculating Alignment Score (AS) at 15~FPS, we replaced the alternate image embeddings $E_V^i$ with their previous ones $E_V^{i-1}$, to ensure it's consistent with the trained model. Similarily, we did for testing at 4~FPS.

\section*{H. Comparison with Baselines.}

We compared our model against 3 different baselines: Diff-Foley, Im2Wav, and Make-an-Audio~\cite{diff_foley, sheffer2023hear, make_an_audio}). Im2Wav samples the input audio at 16~KHz and generates an audio of length 5 seconds. Make-an-Audio also samples at 16~KHz, but generates an audio of length 9-10 seconds with the intermediate spectrogram representation of $80 \times 624$. Diff-Foley also samples at 16~KHz to generate an audio of length 8 seconds via audio spectrogram of shape $256 \times 128$. On the contrary, we samples the input audio waveform at a higher 22~KHz sampling rate and also generates longer audio samples of 10 seconds. Using Stable Diffusion ~\cite{stable_diffusion} allowed us to generate high resolution spectrograms that can fit more frequency bins and longer temporal length. 
For calculating the metrics of baselines, we make sure to use the same length ground truth audio wave as they generate for a fair comparison. We fine-tuned the Im2Wav, Make-an-Audio, and Diff-Foley's CAVP model for a few iterations on our dataset before testing.

\section*{I. More Results}
\begin{figure*}[!ht]
\centering
\includegraphics[width=5in]{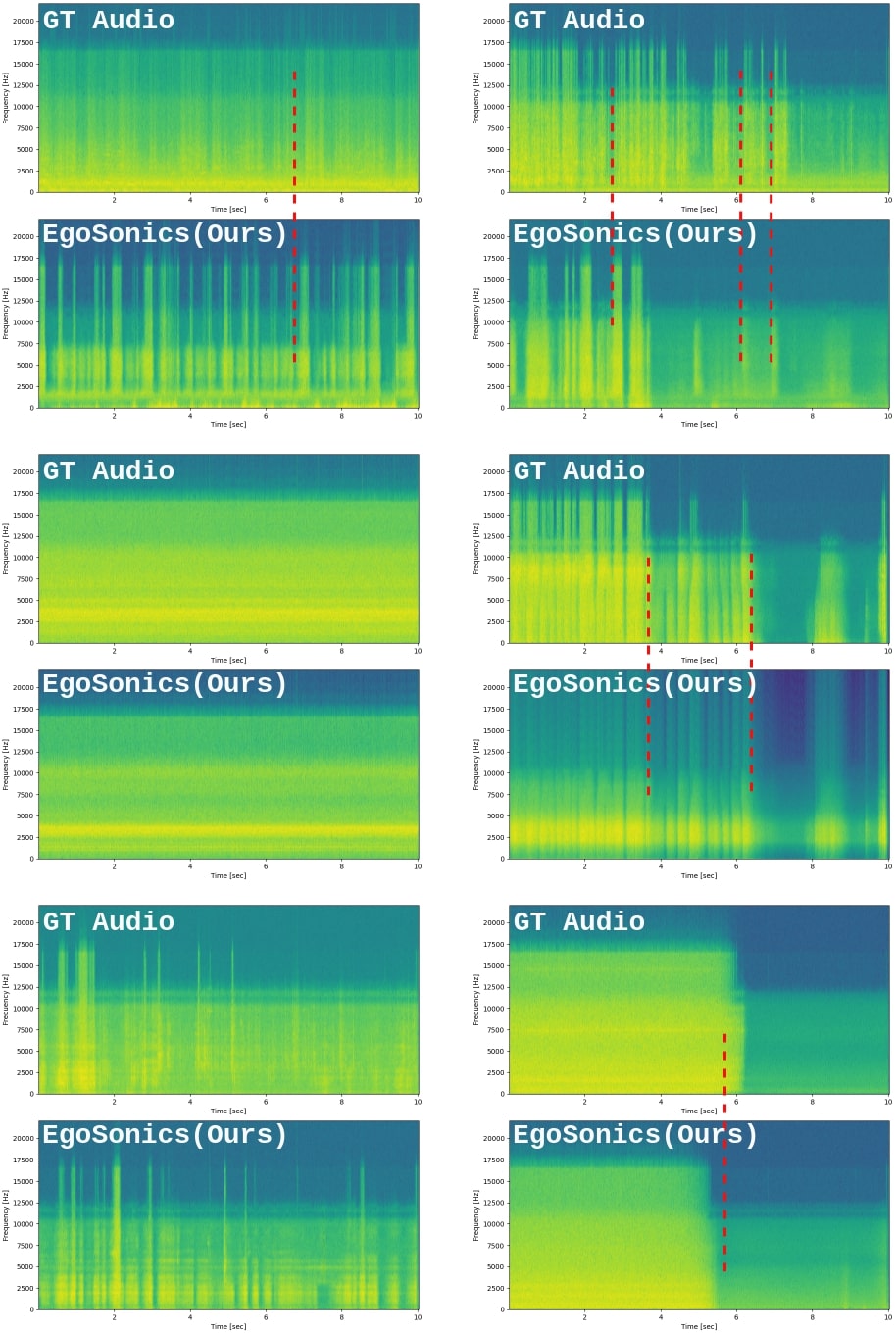} 
\caption{More Results (different color map).}
\label{fig:supp_results}
\end{figure*}

Fig.~\ref{fig:supp_results} shows more results generated from our model. We have used a different color scheme for spectrograms for a different perspective.

\section*{J. Failure Cases, Limitations, and Ethical Considerations}

We also analyzed the failure cases. Most of the failure cases can be classifier into two categories: temporal misalignment and context-level misalignment. The temporal misalignment refers to cases where the model is able to predict contextually meaning audio, however, it's misaligned with the input video. Fig.~\ref{fig:supp_failure}(a) shows some of the misaligned results. The main factor contribution to the misalignment is the lack of rich visual information is most cases. For example, in the first case, a carpenter is polishing a steel bar with a rotating brush and the sound is made when they both are in contact. However, from the video, it's not very clear if the steel bar is actually in contact with the rotating brush or no. 

The other type of failure happens when the model is not able to predict the contextually acceptable audio. This is mainly a reason of lack of data. For example, since there are a very few samples of musical instruments in the Ego4D dataset, our model doesn't perform very well on such videos. The similar thing happens if the model encounters people interacting. We believe that such challenges can be solved by training our model on a large amount of dataset comprising millions of audio-video pairs.

EgoSonics, being a generative model capable of accurately predicting audio from muted video, should be restricted to applications in research, the development of interactive AR/VR technologies, and assistive technologies for individuals with impairments. It is imperative to enforce strict adherence to ethical guidelines to prevent the misuse of this model for unethical purposes.

\end{document}